\documentclass[journal]{vgtc}                

\usepackage[export]{adjustbox}
\usepackage{appendix}
\usepackage{array}
\usepackage{enumitem}
\usepackage{graphicx}
\usepackage{mathptmx}
\usepackage{tikz}
\usepackage{times}

\usepackage{booktabs}

\usepackage[justification=centering]{caption}

\usepackage{float}

\usepackage[none]{hyphenat}
\emergencystretch=\maxdimen
\onlineid{0}

\vgtccategory{Research}

\vgtcinsertpkg

\title{Obstacle Detection for BVLOS Drones\\ \normalsize{NHL Stenden Professorship in Computer Vision \& Data Science}} 

\author{Jan Moros Esteban \\ \small{Supervisors: Maya Aghaei Gavari, Jaap van de Loosdrecht}}
\authorfooter{
\item
  Jan Moros Esteban is a Computing Science student at the NHL Stenden University of Applied Sciences, E-mail: jan.moros.esteban@student.nhlstenden.com.
\item
  Maya Aghaei Gavari is a senior researcher at the NHL Stenden Center of Expertise in Computer Vision \& Data Science, E-mail: maya.aghaei.gavari@nhlstenden.com.
\item
  Jaap van de Loosdrecht is a professor of applied sciences at the NHL Stenden Center of Expertise in Computer Vision \& Data Science, E-mail: jaap.van.de.loosdrecht@nhlstenden.com.
}

\abstract{
With the introduction of new regulations in the European Union, the future of \textit{Beyond Visual Line Of Sight} (BVLOS) drones is set to bloom. This led to the creation of the theBEAST project, which aims to create an autonomous security drone, with focus on those regulations and on safety. This technical paper describes the first steps of a module within this project, which revolves around detecting obstacles so they can be avoided in a fail-safe landing.
A deep learning powered object detection method is the subject of our research, and various experiments are held to maximize its performance, such as comparing various data augmentation techniques or YOLOv3 and YOLOv5. According to the results of the experiments, we conclude that although object detection is a promising approach to resolve this problem, more volume of data is required for potential usage in a real-life application.
} 


\begin{document}

\firstsection{Introduction}

\maketitle

Drones have become widespread in the recent years, as technological advancements have made their components available to the public. They have also been studied as a versatile tool for tasks like search and rescue, transportation or agriculture \cite{europe-drones} .

However, its growth in commercial activities has been hindered, largely by harsh regulations in most European countries. This has been set to change, with the recent publication of new drone regulations for all European Union members. The regulation dated at 11th of June 2019 introduced some clauses which stated that \textit{Beyond Visual Line Of Sight} (BVLOS) drones were allowed in certain situations \cite{eu}.

That change in regulations led to the emergence of the \textit{Towards the first and Best EU-approved Autonomous Security drone for BVLOS flighT} (The BEAST) project, a collaboration among Saxion, NHL Stenden, University of Twente, eight SMEs and five Public partners.

This project is a sub-module of The BEAST, the whole module being \textit{Object avoidance for fail-safe landing}. More specifically, its objective is to detect obstacles on the ground that could cause an unsuccessful landing or people in the frame that must be avoided for their own safety.
 
In flights within the visual line of sight, the operator can view the UAV with their own eyes, and have the supplementary Point of View (PoV) from its camera. For BVLOS, however, the pilot only has the information provided by the UAV's PoV.
Therefore, an autonomous obstacle detection and avoidance system would reduce the risk of crashing in case of malfunction or other unexpected circumstances.

The system must also be self-contained and able to go on board of the drone, to avoid the added risk of making its calculations remotely. This adds a level of complexity to the project, as it needs to be run on a lightweight computer. Fortunately, with the growth in popularity of deep learning, machines specifically designed for running parallel tasks have been introduced \cite{jetson}. 
Although this is a final goal of the module, in this project the main focus will be on optimizing the performance of the project, not its speed and resource consumption.

For this task we chose an object detection approach, to predict an area where the object is located. There are multiple state of the art algorithms for this task, and we chose the YOLO family for their inference speed. More specifically, the models YOLOv3 \cite{yolov3} and YOLOv5 \cite{yolov5} will be compared. \\

This research paper will address object detection for fail-safe landings by answering the main research question: 
\textit{How can an object detection method be used for robust obstacle avoidance in camera-equipped drones?}

To address the main question systematically, three sub-questions have been identified and researched:
\begin{itemize}
    \setlength\itemsep{0.2em}
    \item[-] \textit{To what extent is object detection the appropriate approach for the type of data in this project?}
    \item[-] \textit{How much can the performance of the model increase by applying data augmentation, and which techniques can improve it the most?}
    \item[-] \textit{Which version of the YOLO architecture leads to a better object detection performance towards the goal of this project?}
\end{itemize}

\section{State of the art}

There are several research papers regarding the Detection of Safe Landing Spots for Unmanned Aerial Vehicles (UAV), including drones. These papers include mostly traditional Computer Vision techniques. In their paper from 2006, Bosch et al. \cite{mosaics} use various homographies to estimate a plane which corresponds to the land. Afterwards, the image is thresholded, separating the plane from the obstacles. Moreover, they create a grid fusing all images to make a complete view of the flight. This creates interesting results, at an acceptable 2fps speed, but it has two fundamental flaws: it assumes the ground can be mapped to a plane and that obstacles are stationary.

More recently, Lee et al., on their 2019 paper on UAV landing guidance \cite{s19204468} use a technique based on optical flow. Their approach consists on creating an optical flow magnitude map, created by moving the drone downwards. On that map, obstacles have a higher value than the ground, and thus they can be thresholded. This study shows some promising results, but it relies on manually selected thresholds and needs the drone to make a specific maneuver. \\

As none of the aforementioned studies meets the requisites of our project, we proposed to use deep learning, within the context of either object detection or anomaly detection.

The current state of the art in anomaly detection on images is the 2020 \textit{PANDA} method \cite{panda}. It has a higher ROC AUC than the state of the art over almost all the relevant datasets. Its main characteristic is the use of a pretrained feature extractor, the output of which is used to determine if a sample is an anomaly or not. Our concern with this approach is that our data presents a high number of obstacles, compared to the percentage of anomalies expected by an anomaly detector.

Although anomaly detection remains a sensible path to follow in the later stages of the project when it comes to recognition of unseen obstacles, at this stage we opt for using object detection to perform the anomaly detection. Among the studies on object detection for footage from drones, in \cite{droneOD} authors use a Convolutional Neural Network to create an object detector with various classes. On the other hand, \cite{crowd} uses a Fully Connected Neural Network to extract a heatmap of the probability of a crowd being present. This one diverges from our objective but it can be a valuable source of inspiration.

The main problem with object detection is that it might overfit the dataset and not be able to correctly generalize to different scenarios. Another challenge to be faced is that of the view point. Being captured from a Drone, our dataset consists of images from a bird's eye view. This might mean that most conventional object detection datasets are not suitable for this project. However, some datasets do exist for similar tasks. The aforementioned paper about crowd detection \cite{crowd} uses the VisDrone dataset \cite{zhu2020vision}. This dataset was created as a benchmark, divided in five tasks: image object detection, video object detection, single object tracking, multi-object tracking, and crowd counting. We reckon in the later stages of development it will be beneficial to use the crowd counting part of the dataset to improve the ability of the prototype to detect people in the scene. 

\section{Materials and Methods}

This section describes the materials and methods used throughout the experiments, such as the dataset, software, hardware, and evaluation metrics.

\subsection{Dataset}
A custom dataset was created by the Mechatronics research group at Saxion University. 
It consists of 8 videos captured by their drones. The videos are filmed at two different locations and in various weather conditions, most noticeably after a snow storm. A total of 423 frames was later converted into images, of sizes 3840x2160 or 1920x1080 pixels, depending on the UAV model. 

\begin{figure}[h]
    \centering
	\includegraphics[width=0.3\textwidth]{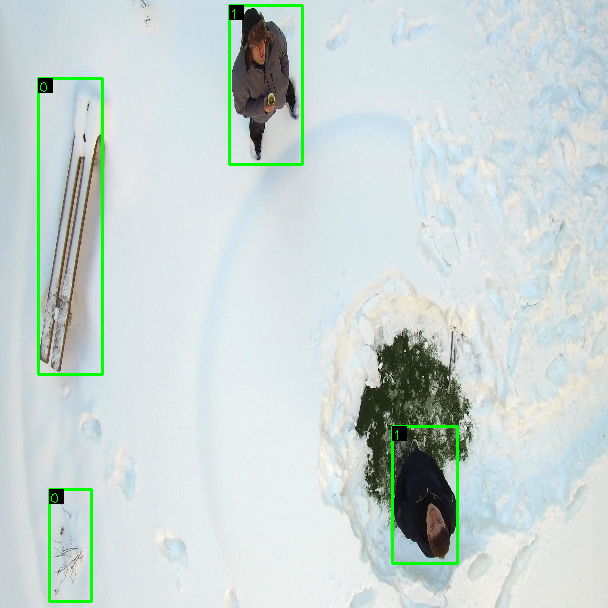} \\
	\vspace{0.4cm}
	\includegraphics[width=0.3\textwidth]{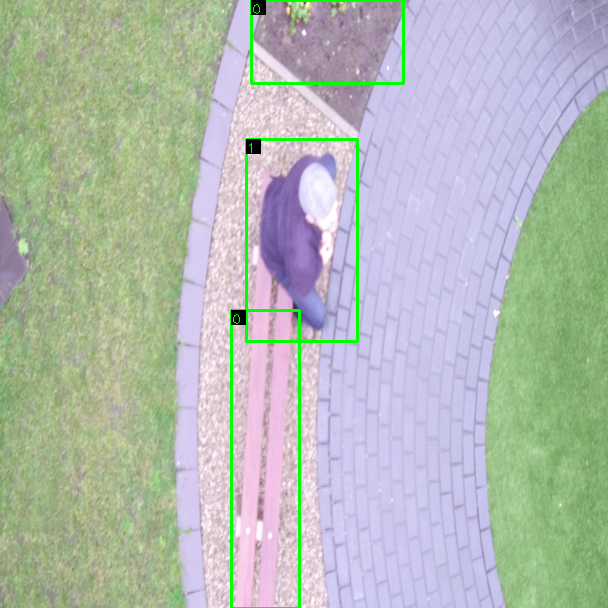}
	\caption{Snowy and cloudy samples from the dataset. \\ Class 0 represents Obstacle and class 1, Person.}
	\label{fig:dataset1}
\end{figure}

Those images were labeled using the LabelMe \cite{labelme} tool. With the topic of the research being object detection, rectangular bounding boxes is the right choice for the label type. As for the classes of the objects to be detected, there are two of them in the annotations: Obstacle and Person. Obstacle refers to any object in the scene that the drone should avoid to land on, and Person is assigned to humans in the scene which must be avoided. Figure \ref{fig:dataset1} shows two samples extracted from the dataset.


The dataset is split in two parts. The first version is used on all experiments unless specified otherwise. Its 6 videos feature the same location filmed by two UAV models and diverse weather and lighting conditions. The second addition consists of two videos. One of them is filmed in the same location as the previous dataset, albeit with different weather. The other one is located at a different scenario, and only contains objects of the class Person.

The first version of the dataset contains 300 images, which were split into 230 training samples, 58 validation samples and 12 testing samples. The new images, which sum to be 123, make the complete dataset of 423 samples. This final version was more evenly split, following a 70/15/15 ratio which results in 296, 64 and 63 images for training, validation and test, respectively.

\subsection{Convolutional Neural Network Architectures}

Currently, Convolutional Neural Networks (CNNs) are the most efficient and commonly used tool for object detection. Many state of the art algorithms exist for this task, but in this paper we will be focusing on the YOLO family, more precisely on its versions v3 \cite{yolov3} and v5 \cite{yolov5}.

YOLOv3 is the third iteration of the open source algorithm You Only Look Once, by Joseph Redmon and Ali Farhadi. Although that affects accuracy, it is able to perform near real-time predictions which can be an advantage for our project. As the name suggests, YOLO uses a single-shot approach, in which it only needs to process each image once. For that, it creates some bounding boxes based on a grid and then applies regression to predict their position and dimensions. Additionally, on version 3, a Feature Pyramid Network is used in order to collect features at three distinct scale levels. This helps the model to be more robust to various object sizes.

YOLOv5 on the other hand, has been developed by the company Ultralytics. It is the first iteration to be fully implemented in PyTorch, without relying on the independent backbone used on the early YOLO architectures, named Darknet. Data augmentation is one of the most notable changes in this version. It applies scaling, color space augmentations and most importantly mosaicing. It does the last one by combining four images into a grid with random ratios.
Another key upgrade is the way it calculates the loss function, as it uses a combination of Generalized Intersection over Union \cite{Rezatofighi_2018_CVPR} and the original YOLO obj and class losses.

Another interesting feature for our project is the fact that YOLOv5 has four model sizes plus a mobile version of each of them, which feature an even more lightweight architecture. These can be used to make a trade-off between speed and accuracy, or to better suit the characteristics of a lightweight computer.



\subsection{Software}
The primary software used is the programming language Python via the Pycharm IDE \cite{pycharm}. Multiple packages were used to add functionalities to it, the most prominent being:

\begin{itemize}
    \item Pytorch \cite{pytorch} for handling the machine learning tasks.
    \item TensorFlow \cite{tensorflow} for its visualization toolkit, TensorBoard.
    \item Albumentations \cite{albumentations} for Data Agumentation.
\end{itemize}

\subsection{Hardware}
Table \ref{tab:hardware} showcases the hardware specifications of the computer used to run the experiments throughout the project.

\begin{table}[h]
\centering
\caption{Hardware specifications}
\label{tab:hardware}
\begin{tabular}{|l|l|}
\hline
\textbf{Hardware} & Virtual Machine hosted at NHL \\ \hline
CPU               & 8 Cores @ 2.4 GHz             \\ \hline
RAM               & 29 GB                         \\ \hline
GPU model         & Nvidia GeForce RTX 2070 SUPER \\ \hline
GPU memory        & 8 GB                          \\ \hline
CUDA version      & 11.0                          \\ \hline
\end{tabular}
\end{table}

\subsection{Transfer Learning}
Transfer learning is a commonly used technique in deep learning. It is based on the premise that models trained for similar tasks tend to focus on the same features. This way, one can use the weights of a previously trained model as a starting point, instead of randomly initializing them. Then, the training process is usually more efficient as the previous model already learned how to extract important features from the samples, and it only needs adjusting to the new task or data.

This technique is used in this research, to improve the quality of the models. 
Both for YOLOv3 and YOLOv5, weights previously trained on the COCO dataset \cite{coco} were downloaded from their respective GitHub repositories \cite{yolov3weights}\cite{yolov5weights}.
COCO is a generic dataset for object detection, from which the model has learned features that define very diverse objects.

\subsection{Data Augmentation}
Given the small size of the initial dataset, we realized data agumentation would be really useful to improve the performance of the models. So, we chose eight augmentation techniques from the Albumentations \cite{albumentations} library to train the model using each of them, one at a time. Those techniques are:

\begin{itemize}
    \item Horizontal flip
    \item Transpose: \textit{As if it was a 2D matrix, switches rows and columns of the image.}
    \item Random rotation of multiples of 90\textdegree
    \item Random shift of RGB channels: \textit{Shifts the values of each color channel slightly.}
    \item Random Hue, Saturation and Value shift: \textit{Converts the image to the HSV format and shifts its channels.}
    \item Random change of Brightness and Contrast
    \item Gamma correction with random gamma between 0.5 and 1.5.
    \item CLAHE histogram equalization
\end{itemize}

\begin{figure}[h]
    \centering
	\includegraphics[width=\columnwidth]{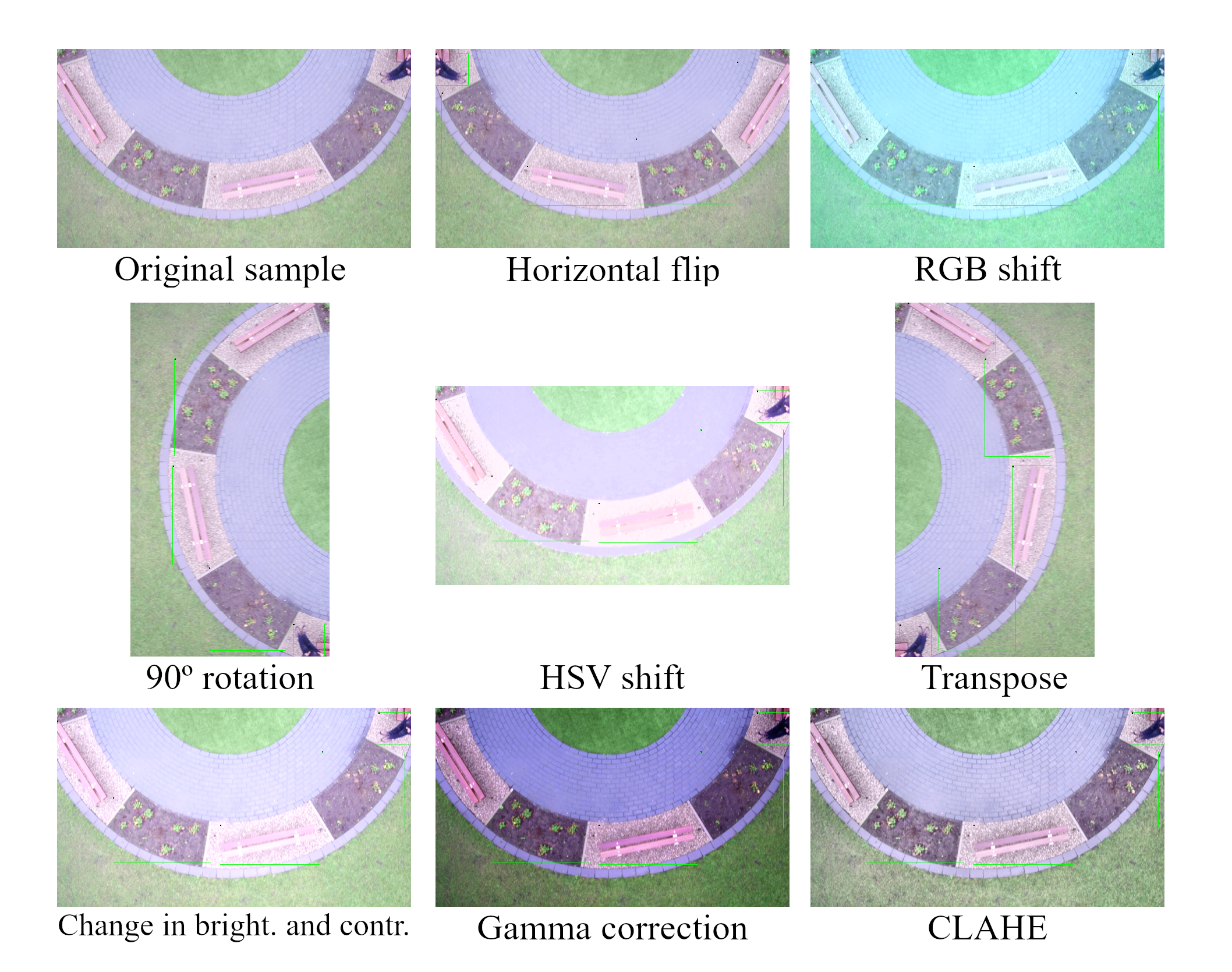}
	\caption{Examples of each augmentation, all applied to the same sample.}
	\label{fig:augmentations}
\end{figure}

They are used with a probability of triggering of 50\% each, on every training sample, and with the default settings from the Albumentations implementation unless stated otherwise in the previous list.

\begin{figure}[h]
    \centering
	\includegraphics[width=\columnwidth]{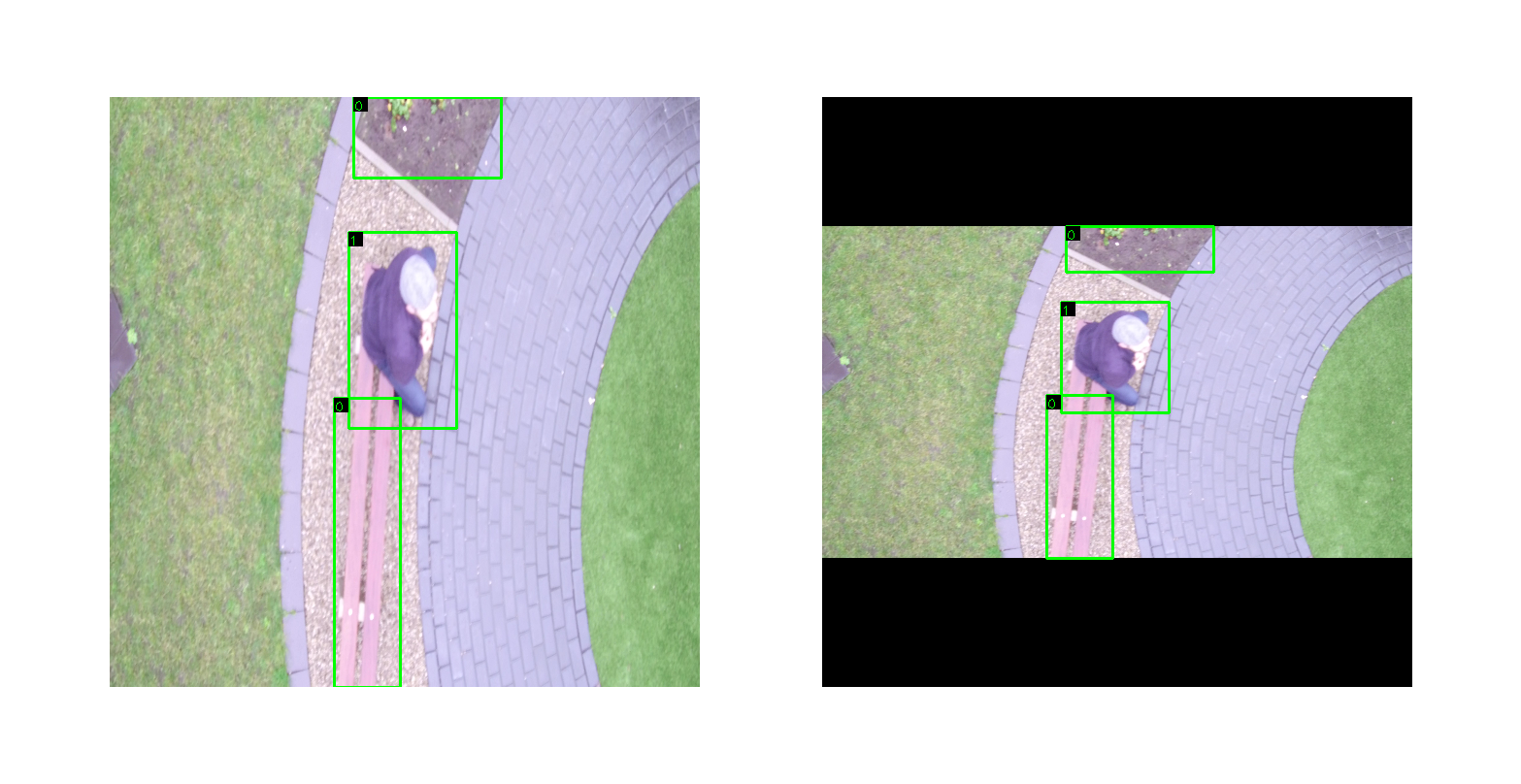}
	\caption{Comparison between the two methods of resizing for the same sample. The lefttmost being stretching, and to its right is letterboxing.}
	\label{fig:letterboxing}
\end{figure}

There is also two different ways to resize the original images so they can be used for training, as seen in Figure \ref{fig:letterboxing}: stretching and letterboxing. The former keeps the image bigger, but the latter adds black stripes where needed to make the image square without losing the original aspect ratio.

\subsection{Evaluation Metrics}
This subsection describes the evaluation metric used by the models to train, and the one used to determine the validity of the predictions.

\subsubsection{Intersection Over Union}
\begin{figure}[h]
    \centering
	\includegraphics[width=0.27\textwidth]{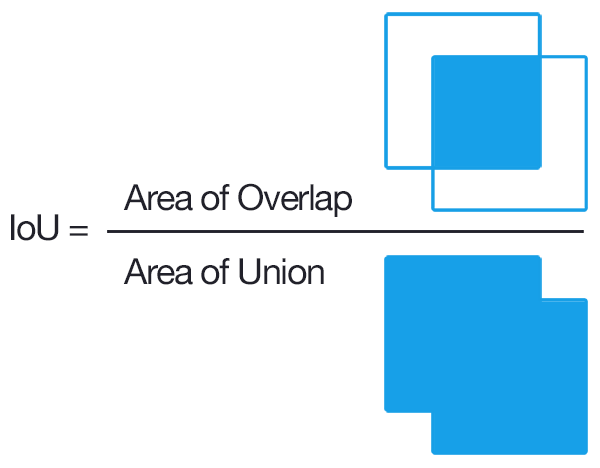}
	\caption{Graphical representation of IoU, by \cite{iou}.}
	\label{fig:iou}
\end{figure}

Internally, both YOLO models use Intersection Over Union (IoU) as a metric to determine the closeness of two bounding boxes. It ranges between 0 and 1, and its calculation can be graphically represented as in Figure \ref{fig:iou} \cite{iou}.

By calculating this ratio, this metric rewards overlapping of the boxes independently from their area.

\subsubsection{Mean Average Precision}
In the context of object detection, mAP refers to the mean of the Average Precision (AP) for all classes. In object detection, IoU is used to calculate the core statistical measures which are needed to calculate AP. A threshold for IoU is chosen, usually 0.5 or 0.95. For this project we used 0.5. The following calculations can then be made.

\begin{itemize}
    \item True Positives = Bounding Boxes (BBoxes) where IoU with Ground Truth (GT) $>$ threshold.
    \item False Positives = BBoxes where IoU with GT $<$ threshold \textit{or} BBoxes that match an already matched GT box.
    \item False Negatives = GT boxes without an assigned BBox.
\end{itemize}

We can now calculate Precision and Recall, using the traditional definitions, as seen in Equation 1.

\begin{equation}
\label{eqn:pr}
Precision = \frac{TP}{TP + FP} \ \ \ \ \ \ \ \ \ \
\ \ Recall = \frac{TP}{TP + FN} \
\end{equation}

Then, the BBoxes are ordered by their class confidence (from the model), and Precision and Recall are calculated for each of them. A Precision-Recall curve can be interpolated from those results. In our implementation, the curve consists of 11 points.

Finally, the mean Average Precision for each image is equal to the area under the Precision-Recall curve. This results in a number between 0 and 1, or 0 to 100\%, in which higher values indicate better performance.

\section{Experiments and Results}
This section aims to describe the experiments that were performed to answer the research questions of this project, as well as showcasing their results.

\subsection{Baseline definition}

In the first experiment, we calculate the mAP score for a YOLOv3 model on the test set when trained with randomly initialized weights, input images stretched to 608x608 pixels and without any augmentation.

We then apply the transfer learning technique to another model, using the weights described in the Materials and Methods section. Its resulting mAP will be used as a baseline for later experiments, as they will be built on top of a model trained with this technique. 

\begin{table}[H]
\caption{Comparison in the model's mAP score between training with randomized or pretrained weights.}
\label{tab:v3raw}
\centering
\renewcommand{\arraystretch}{1.2}
\begin{tabular}{lcc}
\toprule
\textbf{Model}  & \textbf{Randomized} & \textbf{Pretrained}\\ \midrule
YOLOv3 & 58.14 & \textbf{79.12}\\ \bottomrule
\end{tabular}
\end{table}

\begin{figure}[h]
    \centering
	\includegraphics[width=0.85\columnwidth]{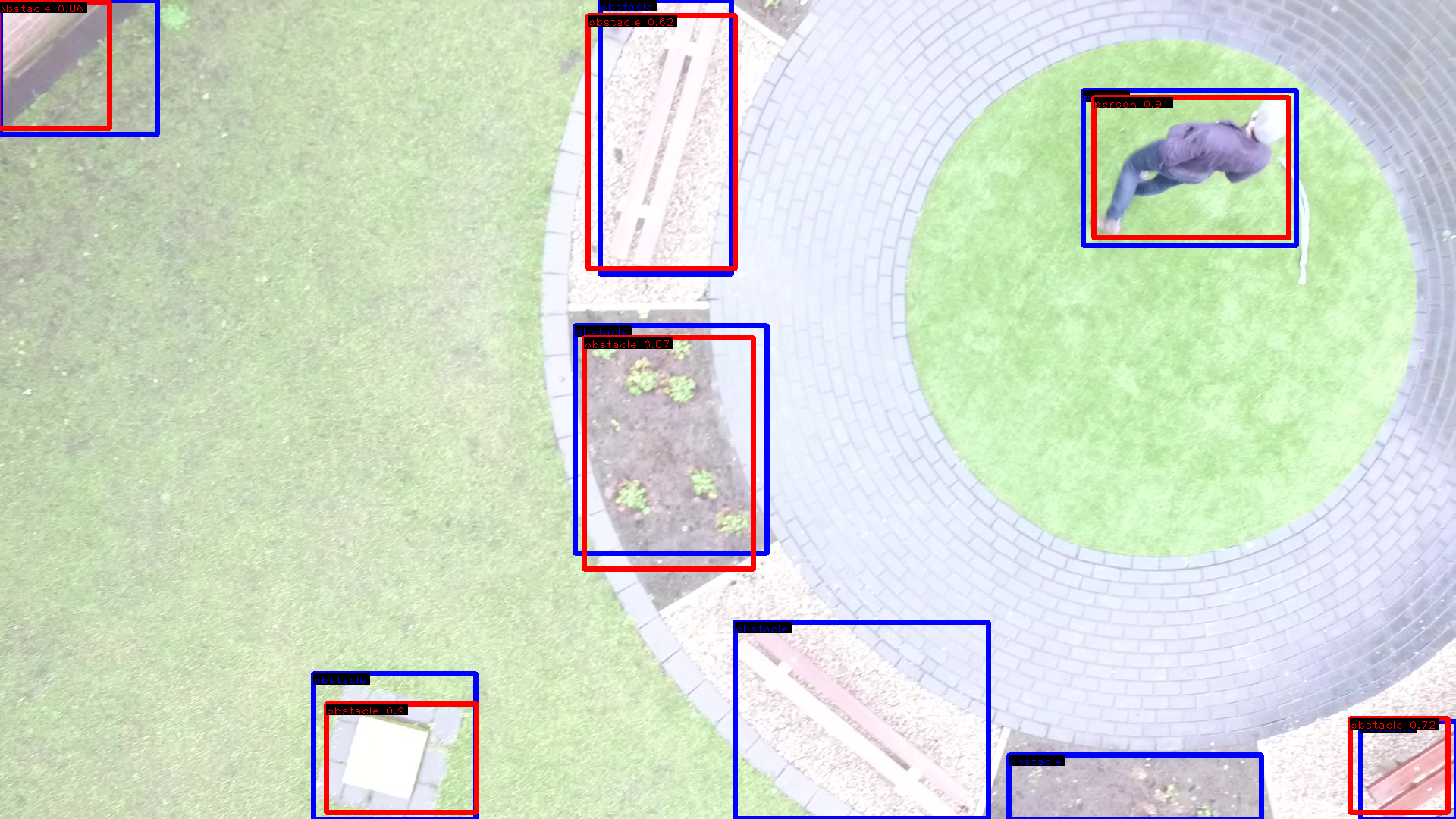}
	\caption{Sample output from the baseline model. Blue rectangles represent the groundtruth detection performed by a human, and red ones represent the output of our model.}
	\label{fig:v3transfer}
\end{figure}

As shown in Table \ref{tab:v3raw}, the model which uses pretrained weight gets a 20\% increase in performance. The almost 80\% performance score can also be graphically visualized in Figure \ref{fig:v3transfer}.

We can see how the algorithm is capable of detecting the Person in the scene, as well as most Obstacles, but misses the bench and ground patch in the bottom center.

\subsection{Data Augmentation and Resize method}
\label{section:data_augm}
With this starting point, we designed the main experiments. The idea is to reach a satisfactory state with YOLOv3, and eventually use its best performing settings to compare it with YOLOv5.

Accordingly, Experiment 2 focuses on the effect of data augmentation on YOLOv3. Its objective is to find which augmentations improve the performance of the model for our current dataset. Simultaneously, the resizing methods are also analyzed, as all data augmentation techniques are applied twice, once with stretching and the other with letterboxing.

The results of Experiment 2 are presented in Table \ref{tab:letterboxing}, where we compare the mean and standard deviation of the mAP score for each method of resizing and augmentation. The mean is calculated from a set of 2 to 4 launches, to make up for the variance caused by the randomness of the Dataset.

\begin{table}[h]
\caption{Comparison in mean (std) mAP score between the different augmentations and resize methods.}
\label{tab:letterboxing}
\setlength{\tabcolsep}{0.4\tabcolsep}
\renewcommand{\arraystretch}{1.2}
\centering
\resizebox{\columnwidth}{!}{%
\begin{tabular}{l c c}
\toprule
& \multicolumn{2}{c}{\textbf{Resize method}} \\
\cmidrule{2-3}
\textbf{Augmentation}    & \multicolumn{1}{l}{Stretching} & \multicolumn{1}{l}{Letterboxing} \\
\midrule[0.6pt]
\textit{Without augmentations}           & \textit{79.12 (6.41)}                 & \textbf{\textit{88.10 (0.00)}}                 \\ \midrule[0.1pt]
Horizontal flip                          & \underline{81.64 (3.71) }                         & \underline{\textbf{89.15 (0.27)}}                  \\
Image matrix transpose                   & 78.41 (6.82)                          & \textbf{86.12 (6.55)}              \\
Random rotation of multiples of 90°      & \underline{85.23 (4.35)}                          & \underline{\textbf{90.00 (0.37)}}              \\
Random Hue, Saturation and Value shift   & 72.34 (11.65)                         & \textbf{76.19 (0.01)}                 \\
Random change of Brightness and Contrast & \underline{\textbf{84.73 (7.14)}}                          & 77.36 (4.07)                          \\
Gamma correction with random gamma       & \underline{\textbf{81.55 (0.46)}}                          & 81.52 (3.39)                 \\
Random shift of RGB channels             & 71.11 (14.03)                         & \textbf{78.84 (2.46)}                          \\
CLAHE histogram equalization             & \underline{80.11 (5.41)}                          & \textbf{86.14 (2.31)}     \\
\bottomrule
\end{tabular}
}
\end{table}

Underlined cells in Table \ref{tab:letterboxing} highlight which augmentations a higher performance than the same model without augmentation. 

An interesting insight from the Standard Deviation can be seen in the two color shift augmentations when the stretching method is used. They have much higher variance in their score, which is not desirable when trying to achieve the highest performance possible.

As shown by the bold cells, Letterboxing achieves a higher score than Stretching in most augmentations. This confirms our hypothesis, as it is expected to perform better in object detection algorithms. This is due to it preserving the original aspect ratio of the image, which is heavily distorted by stretching.

\subsection{YOLOv5 sizes}
\label{section:sizes}

Moving to YOLOv5, Experiment 3 makes a comparison between the architecture's different sizes. For those we chose the Small, Medium and Large.

\begin{table}[H]
\caption{Comparison in mAP score between various sizes of the YOLOv5 architecture.}
\label{tab:sizes}
\renewcommand{\arraystretch}{1.2}
\centering
\begin{tabular}{lcc}
\toprule
\textbf{YOLOv5 Size}  & \textbf{Mean}   & \textbf{Std}     \\ \midrule
Small                 & 87.59          & 2.24 \\
Medium                & 89.68          & 0.77 \\
Large                 & \textbf{92.16}          & 1.99 \\ \bottomrule
\end{tabular}
\end{table}

Table \ref{tab:sizes} shows that bigger sizes lead to higher mAP score. This asserts the analysis of its authors \cite{yolov5}. On the other hand, it contradicts our initial hypothesis which was that for such a tiny dataset the small version would perform better.

In spite of these results, we used the small version of the model through the rest of the research because of the characteristics of the project. As it has to be run in real time on a lightweight computer, we believe the small model can be a better option due to its prediction speed and its smaller file size. However, more research needs to be done in order to determine which model is the hardware able to run properly. Depending on the power of the computer, a YOLOv5m model could also be a viable alternative.

\subsection{YOLOv5 baseline}
With the final choice of size being YOLOv5s, Experiment 4 repeats the same process of Experiment 1 on YOLOv5, to establish a baseline.

\begin{table}[H]
\caption{Comparison in the model's mAP score between training with randomized or pretrained weights.}
\label{tab:raw}
\centering
\renewcommand{\arraystretch}{1.2}
\begin{tabular}{lcc}
\toprule
\textbf{Model}  & \textbf{Randomized} & \textbf{Pretrained}\\ \midrule
YOLOv5s & 80.90 & \textbf{87.59}\\ \bottomrule
\end{tabular}
\end{table}

Again, we can see in Table \ref{tab:raw} that the transfer learning technique improves the model's performance. It is a milder improvement than the one on YOLOv3, but sets the mAP score to almost 90\%.

\subsection{Augmentations combination}

Experiment 5 is a comparison between YOLOv3 and YOLOv5s, using the best resize method and augmentations found in previous experiments.

Due to time limitations, the top four augmentations using the stretching method were chosen. Those are Horizontal flip, Rotation of multiple of 90º, Random change in brightness and contrast and Gamma correction.

As the resize method, Letterboxing was chosen as it proved to be superior in Experiment 2.

\begin{table}[H]
\caption{Comparison in mAP score between the two models, both trained with the combination of augmentations.}
\label{tab:combo}
\centering
\renewcommand{\arraystretch}{1.2}
\begin{tabular}{lcc}
\toprule
\textbf{Model} & \textbf{Mean}   & \textbf{Std}     \\ \midrule
YOLOv3 & 84.69 & 2.04 \\
YOLOv5s & \textbf{92.42} & 2.14 \\ \bottomrule
\end{tabular}
\end{table}

As clearly visible in Table \ref{tab:combo}, YOLOv5 outperforms its predecessor, achieving an 8\% higher mean mAP score with similar deviation. Also, the combination of augmentations is an improvement over the unaugmented YOLOv5 model, with almost a 5\% increase in mAP.

\begin{figure}[h]
    \centering
	\includegraphics[width=0.85\columnwidth]{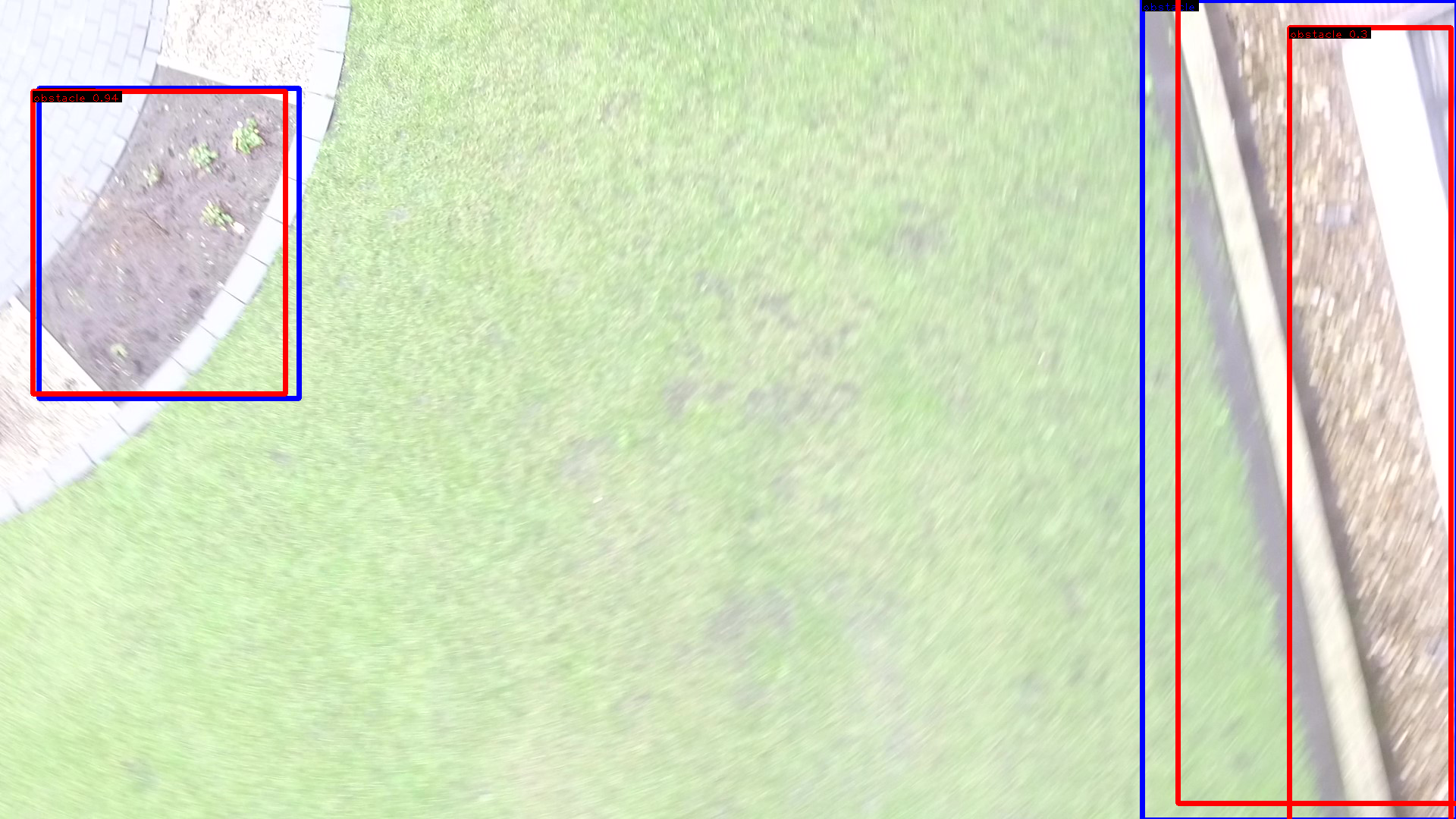}
	\caption{Sample output from the best model with an erroneous prediction.}
	\label{fig:v5_combo_wrong}
\end{figure}

Figure \ref{fig:v5_combo_wrong} showcases the main reason the best performing model does not achieve an even higher mAP score. Big patches of ground like the one in the right of the image are sparse in the dataset, which makes it hard for the model to learn their features. As in all object detectors, some variations in the coordinates of the predicted box prevent achieving a perfect score.

\subsection{Dataset Size}
\label{section:dataset_size}
The final two experiments regard the second part of the dataset. For experiment 6, the best model trained with the first version of the dataset is tested on all of the new images from version 2, to observe how it generalizes. For Experiment 7, the best settings found in Experiments 3 and 4 are used to train a new model using the complete version of the dataset.

\begin{table}[H]
\caption{mAP Score of the best YOLOv5s model when trained with the full dataset, and when testing on the new data.}
\label{tab:dataset}
\centering
\renewcommand{\arraystretch}{1.2}
\begin{tabular}{lc}
\toprule
\textbf{Experiment}                       & \textbf{mAP Score} \\ \midrule[0.6pt]
Only test with new data          & 46.34    \\ \midrule[0.1pt]
Train and test with full dataset & 90.19    \\ \bottomrule
\end{tabular}
\end{table}

\begin{figure}[h]
    \centering
	\includegraphics[width=0.7\columnwidth]{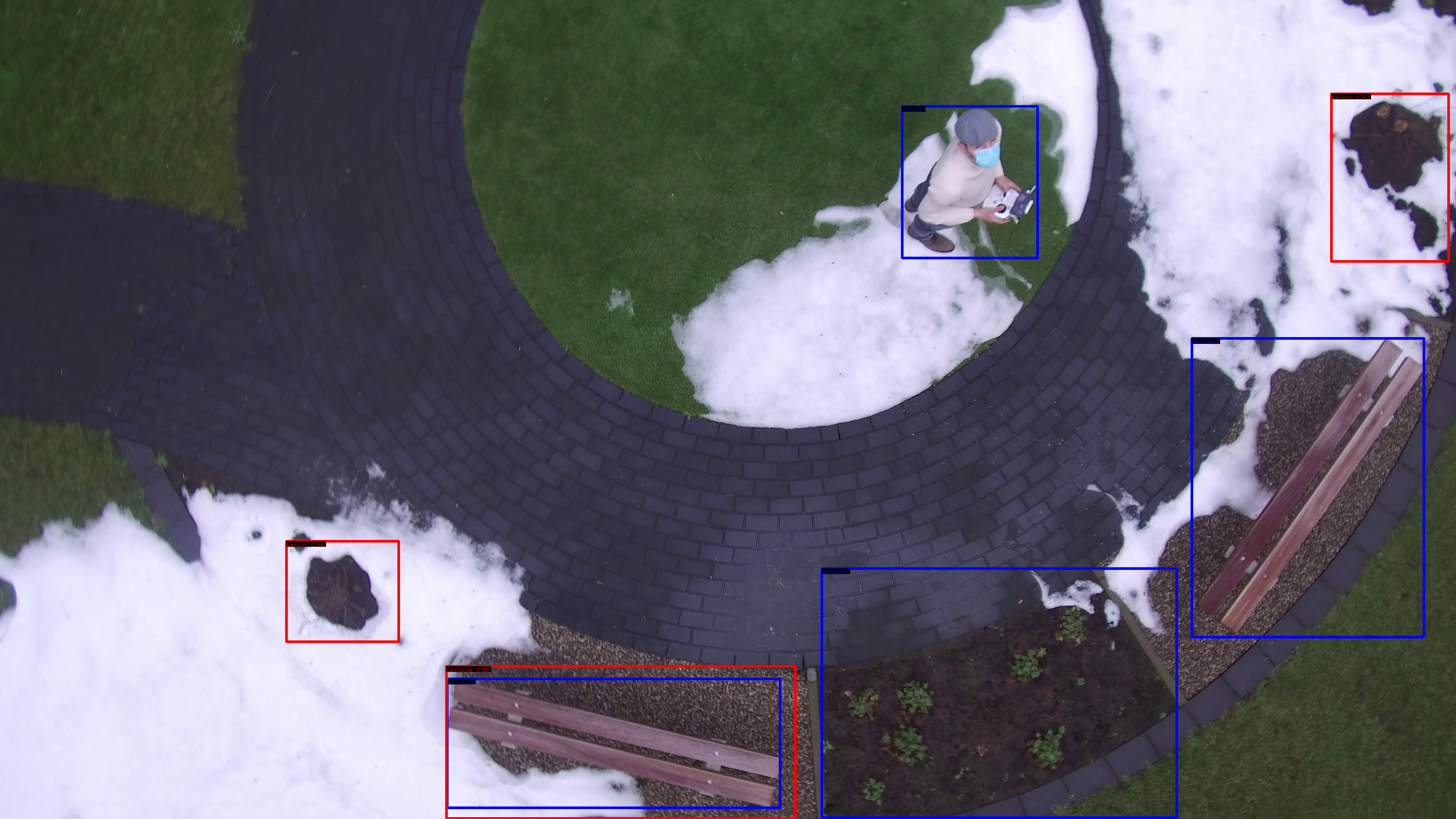} \\ 
	\vspace{0.5cm}
	\includegraphics[width=0.7\columnwidth]{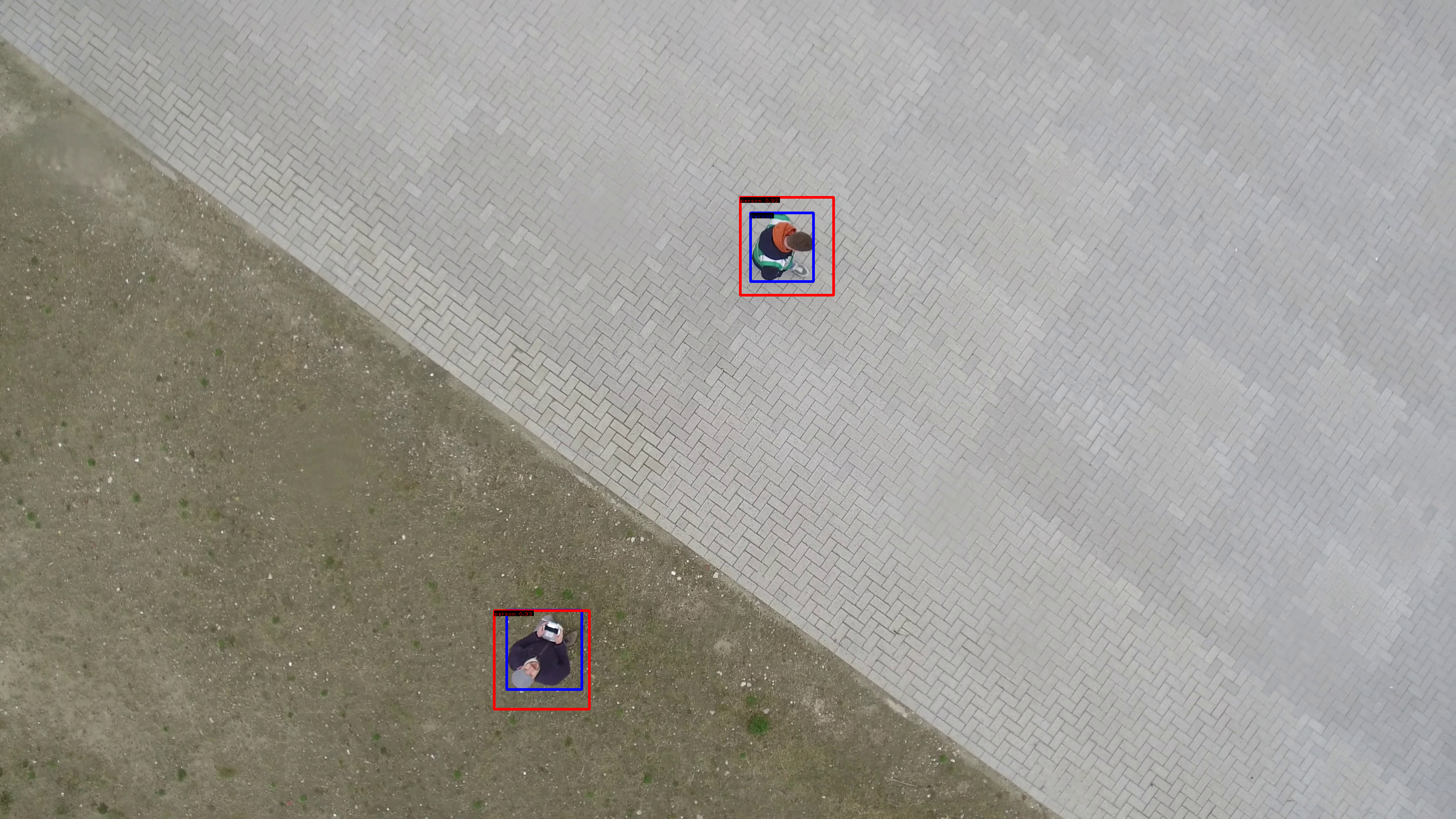}
	\caption{Output samples of the model from Experiment 3 on the new dataset.}
	\label{fig:full_test}
\end{figure}

The first one, designed to analyze the generalization ability of our model, gets an mAP score of 46\%. To our opinion, this is a logical result. As visible in Figure \ref{fig:full_test}a, the model struggles with the objects that are different from the original dataset, such as the bushes which it labels as Person. On the other hand, in Figure \ref{fig:full_test}b it correctly detects the two people in the scene, even if they are in an unseen environment. This last occurrence might be affected by the weights pretrained on COCO, as discussed in subsection.

The second one was just to feed more data into the training, so the overall performance in real situations improves. It gives satisfactory results, as it achieves a 90\% mAP score. There is a slight decline of 2\% with respect to the same experiment with the small dataset, which could as well be due to the change in the test set size.

\section{Discussion, Conclusion \& Future Work}

This section sums up the paper by describing the discussion, conclusion and future work of this research.

\subsection{Discussion}

Going back to the Research Question, while based on the limited available dataset, we can conclude that a deep learning method based on object detection is a suitable approach to detect obstacles from a drone camera. 

About the second sub-question, as showcased in section \ref{section:data_augm}, data augmentation has proven to increase the performance of the model. Moreover, a number of techniques have been selected for the performance boost they provide to the model. Those are Horizontal flip, Rotation of multiple of  90º, Random  change in brightness and contrast and Gamma correction. CLAHE also improved over the baseline, but was cut off to reduce the number of augmentations used.

The difference between YOLOv3 and YOLOv5 in terms of performance has also been discussed in multiple experiments, towards answering sub-question number three.
The results of Experiment 4, represented in Table \ref{tab:raw}, together with Table \ref{tab:v3raw} can be used as the first comparison between the two architectures.

The contrast confirms our hypothesis: On the one hand, YOLOv5s outperforms YOLOv3 in both experiments. On the other hand, both models increase their performance significantly by using transfer learning. This is also as envisioned, because training starts with prior knowledge instead of a randomized initialization.

Another interesting reflection on the results of Experiments 3 and 6 lays on the fact that the models use weights pretrained on the COCO dataset. Benches being the most common Obstacle and Person the other class, it is possible that the pretrained model already knows the features that define those two objects. 

As bigger models can store more information about those features, it might be a reason behind the increase in performance seen in section \ref{section:sizes}, even with such a small dataset. This might as well be the case for the first experiment in section \ref{section:dataset_size}, as it is able to detect persons more easily than obstacles. This could be caused by the large quantity of Human objects in the COCO dataset \cite{coco}.

Nevertheless, this research has a big limitation. Due to the small and limited dataset that was available, it is not possible to make good conclusions for real world scenarios. This renders most conclusions biased, as they do not represent the performance of the problem in a real scenario. 

\subsection{Conclusion}

In its current state, the product of this project cannot yet be used in a real life application due to the small size of the dataset, but it is a first step towards a safe-landing mechanism which can be really helpful as a security measure in a future society, if the use of drones keeps rising at the rate it is predicted.

This conclusion made us take the decision to shift the focus of the project away from performing extensive research. The definitive contribution of this project is that a data processing pipeline was created and tested, and that it can be used to assess the quality of the methods used when the partners in the project have a comprehensive dataset available.

In this path, the project counts with two complete train, validation and test pipelines, one for each architecture. Also, they are integrated with support for Albumentations data augmentation, both methods of image resizing shown in the paper, ease to use custom configuration and weight files and various tools to store and showcase training progress and results.

Nonetheless, the conclusions from the research can still be used in such later development, as we have found which augmentations work best for this architectures and type of data, and most importantly that object detection algorithms such as YOLOv5 are the way to proceed with obstacle detection.

\subsection{Future Work}

The immediate next step for this module of the theBEAST project is to gather more varied data. What characterizes deep learning is the way it learns to generalize, but it needs a solid base to be able to do so.

An addition to it might be the crowd counting dataset from VisDrone \cite{zhu2020vision}, as mentioned in the State of the art section. It could be used to improve the model's robustness with the Person class, and to train it to be able to detect larger amounts of them in the same frame.

After that, the main experiments created during the course of this research can be repeated, to test their validity and choose the configuration that maximizes performance for the new data as well.

Alternatively, an approach based on anomaly detection for images could be tried. This is a less developed field than object detection, so it may be hard to find good architectures that are able to get good results at the speed the drone needs them, but it can be worth to try.

\acknowledgements{
	We thank Gabriel Damian (Saxion University of Applied Sciences) for his contribution to the research, and for constructing the dataset used throughout its development. \\
	
	\noindent This project is financially supported by the RAAK-mkb program.
}

\bibliographystyle{unsrt}
\bibliography{template}

\begin{thebibliography}{10}

\bibitem{europe-drones}
{SESAR Joint Undertaking}.
\newblock {European Drones Outlook Study, Unlocking the value for Europe}.
\newblock 2016.
\newblock {https://www.sesarju.eu/sites/default/files/
  documents/reports/European\_Drones\_Outlook\_Study\_2016.pdf}.

\bibitem{eu}
{COMMISSION IMPLEMENTING REGULATION (EU)} 2019/947 of 24 {M}ay 2019 on the
  rules and procedures for the operation of unmanned aircraft.
\newblock {\em OJ}, L 152:45, 2019-06-11.

\bibitem{jetson}
{NVIDIA}.
\newblock {Jetson Nano: Deep Learning Inference Benchmarks}.
\newblock {https://developer.nvidia.com/embedded/jetson-nano-dl-
  inference-benchmarks}, n.d.
\newblock Accessed: 2021-02-18.

\bibitem{yolov3}
Joseph Redmon and Ali Farhadi.
\newblock Yolov3: An incremental improvement, 2018.
\newblock arXiv:1804.02767.

\bibitem{yolov5}
Glenn Jocher, Alex Stoken, Jirka Borovec, NanoCode012, ChristopherSTAN, Liu
  Changyu, Laughing, tkianai, yxNONG, Adam Hogan, lorenzomammana, AlexWang1900,
  Ayush Chaurasia, Laurentiu Diaconu, Marc, wanghaoyang0106, ml5ah, Doug,
  Durgesh, Francisco Ingham, Frederik, Guilhen, Adrien Colmagro, Hu~Ye,
  Jacobsolawetz, Jake Poznanski, Jiacong Fang, Junghoon Kim, Khiem Doan, and
  Lijun Yu.
\newblock {ultralytics/yolov5: v4.0 - nn.SiLU() activations, Weights \& Biases
  logging, PyTorch Hub integration}, January 2021.

\bibitem{mosaics}
Sébastien Bosch, Simon Lacroix, and Fernando Caballero.
\newblock {Autonomous Detection of Safe Landing Areas for an UAV from Monocular
  Images}.
\newblock In {\em {IEEE International Conference on Intelligent Robots and
  Systems}}, pages 5522 -- 5527, 2006.
\newblock {https://doi.org/10.1109/IROS.2006.282188}.

\bibitem{s19204468}
Joon~Yeop Lee, Albert~Y. Chung, Hooyeop Shim, Changhwan Joe, Seongjoon Park,
  and Hwangnam Kim.
\newblock Uav flight and landing guidance system for emergency situations †.
\newblock {\em Sensors}, 19(20):Article ID 4468, 2019.
\newblock {https://doi.org/10.3390/s19204468}.

\bibitem{panda}
Tal Reiss, Niv Cohen, Liron Bergman, and Yedid Hoshen.
\newblock {PANDA -- Adapting Pretrained Features for Anomaly Detection}.
\newblock 2020.
\newblock arXiv:2010.05903v1.

\bibitem{droneOD}
Chenfan Sun, Wei Zhan, Jinhiu She, and Yangyang Zhang.
\newblock Object detection from the video taken by drone via convolutional
  neural networks.
\newblock {\em Mathematical Problems in Engineering}, vol. 2020:Article ID
  4013647, 2020.
\newblock {https://doi.org/10.1155/2020/4013647}.

\bibitem{crowd}
Giovanna Castellano, Ciro Castiello, Corrado Mencar, and Gennaro Vessio.
\newblock {Crowd Detection for Drone Safe Landing Through Fully-Convolutional
  Neural Networks}.
\newblock In {\em {SOFSEM 2020: Theory and Practice of Computer Science}},
  pages 301--312, 2020.
\newblock {https://doi.org/10.1007/978-3-030-38919-2\_25}.

\bibitem{zhu2020vision}
Pengfei Zhu, Longyin Wen, Dawei Du, Xiao Bian, Qinghua Hu, and Haibin Ling.
\newblock Vision meets drones: Past, present and future.
\newblock {\em arXiv preprint arXiv:2001.06303}, 2020.

\bibitem{labelme}
Kentaro Wada.
\newblock {labelme: Image Polygonal Annotation with Python}.
\newblock {https://github.com/wkentaro/labelme}, 2016.

\bibitem{Rezatofighi_2018_CVPR}
Hamid Rezatofighi, Nathan Tsoi, JunYoung Gwak, Amir Sadeghian, Ian Reid, and
  Silvio Savarese.
\newblock Generalized intersection over union.
\newblock June 2019.

\bibitem{pycharm}
{Jetbrains}.
\newblock {Pycharm}.
\newblock {https://www.jetbrains.com/pycharm/}, n.d.
\newblock Accessed: 2021-04-04.

\bibitem{pytorch}
Adam Paszke, Sam Gross, Francisco Massa, Adam Lerer, James Bradbury, Gregory
  Chanan, Trevor Killeen, Zeming Lin, Natalia Gimelshein, Luca Antiga, Alban
  Desmaison, Andreas Kopf, Edward Yang, Zachary DeVito, Martin Raison, Alykhan
  Tejani, Sasank Chilamkurthy, Benoit Steiner, Lu~Fang, Junjie Bai, and Soumith
  Chintala.
\newblock Pytorch: An imperative style, high-performance deep learning library.
\newblock In H.~Wallach, H.~Larochelle, A.~Beygelzimer, F.~d\textquotesingle
  Alch\'{e}-Buc, E.~Fox, and R.~Garnett, editors, {\em Advances in Neural
  Information Processing Systems 32}, pages 8024--8035. Curran Associates,
  Inc., 2019.

\bibitem{tensorflow}
Mart\'{\i}n Abadi, Ashish Agarwal, Paul Barham, Eugene Brevdo, Zhifeng Chen,
  Craig Citro, Greg~S. Corrado, Andy Davis, Jeffrey Dean, Matthieu Devin,
  Sanjay Ghemawat, Ian Goodfellow, Andrew Harp, Geoffrey Irving, Michael Isard,
  Yangqing Jia, Rafal Jozefowicz, Lukasz Kaiser, Manjunath Kudlur, Josh
  Levenberg, Dandelion Man\'{e}, Rajat Monga, Sherry Moore, Derek Murray, Chris
  Olah, Mike Schuster, Jonathon Shlens, Benoit Steiner, Ilya Sutskever, Kunal
  Talwar, Paul Tucker, Vincent Vanhoucke, Vijay Vasudevan, Fernanda Vi\'{e}gas,
  Oriol Vinyals, Pete Warden, Martin Wattenberg, Martin Wicke, Yuan Yu, and
  Xiaoqiang Zheng.
\newblock {TensorFlow}: Large-scale machine learning on heterogeneous systems,
  2015.
\newblock Software available from tensorflow.org.

\bibitem{albumentations}
Alexander Buslaev, Vladimir~I. Iglovikov, Eugene Khvedchenya, Alex Parinov,
  Mikhail Druzhinin, and Alexandr~A. Kalinin.
\newblock Albumentations: Fast and flexible image augmentations.
\newblock {\em Information}, 11(2), 2020.

\bibitem{coco}
Tsung-Yi Lin, Michael Maire, Serge Belongie, Lubomir Bourdev, Ross Girshick,
  James Hays, Pietro Perona, Deva Ramanan, C.~Lawrence Zitnick, and Piotr
  Dollár.
\newblock Microsoft coco: Common objects in context, 2014.
\newblock arxiv:1405.0312.

\bibitem{yolov3weights}
{Ultralytics}.
\newblock {ultralytics / yolov3}.
\newblock {https://github.com/ultralytics/yolov3/ tree/archive}, 2020.
\newblock Accessed: 2021-04-04.

\bibitem{yolov5weights}
{Ultralytics}.
\newblock {ultralytics / yolov5}.
\newblock {https://github.com/ultralytics/yolov5}, 2021.
\newblock Accessed: 2021-05-26.

\bibitem{iou}
{Adrian Rosebrock}.
\newblock {Intersection over Union (IoU) for object detection}.
\newblock {https://www.pyimagesearch.com/2016/11/07/intersection-over-union-iou
  -for-object-detection/}, November 2016.
\newblock Accessed: 2021-04-04.

\end{thebibliography}


\end{document}